\documentclass[letterpaper, 10 pt, conference]{ieeeconf}  

\IEEEoverridecommandlockouts                              

\usepackage{xcolor}
\usepackage{graphicx} 
\usepackage{amsmath} 
\graphicspath{{./images/}}

\usepackage{fancyhdr}
\title{\LARGE \bf
Look No Deeper: Recognizing Places from Opposing Viewpoints under Varying Scene Appearance using Single-View Depth Estimation
}

\author{Sourav Garg$^{1}$, Madhu Babu V$^{2}$, Thanuja Dharmasiri$^{3}$, Stephen Hausler$^{1}$,\\ Niko Suenderhauf$^{1}$, Swagat Kumar$^{2}$, Tom Drummond$^{3}$, Michael Milford$^{1}$
\thanks{$^{1}$Australian Centre for Robotic Vision, Queensland University of Technology (QUT), Brisbane, Australia.}%
\thanks{$^{2}$TATA Consultancy Services, Bangalore, India.}%
\thanks{$^{3}$Australian Centre for Robotic Vision, Monash University, Melbourne, Australia.}%
}

\begin{document}

\maketitle
\thispagestyle{fancy}
\pagestyle{empty}

\begin{abstract}
Visual place recognition (VPR) - the act of recognizing a familiar visual place - becomes difficult when there is extreme environmental appearance change or viewpoint change. Particularly challenging is the scenario where both phenomena occur simultaneously, such as when returning for the first time along a road at night that was previously traversed during the day in the opposite direction. While such problems can be solved with panoramic sensors, humans solve this problem regularly with limited field of view vision and without needing to constantly turn around. In this paper, we present a new depth- and temporal-aware visual place recognition system that solves the opposing viewpoint, extreme appearance-change visual place recognition problem. Our system performs \textit{sequence-to-single} matching by extracting depth-filtered keypoints using a state-of-the-art depth estimation pipeline, constructing a keypoint sequence over multiple frames from the reference dataset, and comparing those keypoints to those in a single query image. We evaluate the system on a challenging benchmark dataset and show that it consistently outperforms state-of-the-art techniques. We also develop a range of diagnostic simulation experiments that characterize the contribution of depth-filtered keypoint sequences with respect to key domain parameters including degree of appearance change and camera motion.
\end{abstract}


\section{INTRODUCTION}
Visual Place Recognition (VPR) is a widely researched topic, with recent approaches benefiting from modern deep-learning techniques~\cite{sunderhauf2015place,vysotska2016lazy,chen2017only,arandjelovic2016netvlad,noh2017large}, especially for performance under challenging appearance variations. However, most of the existing literature dealing with such extreme appearance variations has only considered limited changes in viewpoint~\cite{milford2012seqslam,naseer2014robust}. In this paper, we address the more difficult problem of visual place recognition from opposite viewpoints under extreme appearance variations. Our system extracts depth-filtered keypoints using a state-of-the-art depth estimation pipeline ~\cite{babu2018deeper,dharmasiri2018eng,ummenhofer2017demon}, constructing a keypoint sequence over multiple frames from the reference dataset, and compares those keypoints to those in a single query image.

\begin{figure}
\centering
 \includegraphics[scale=0.3]{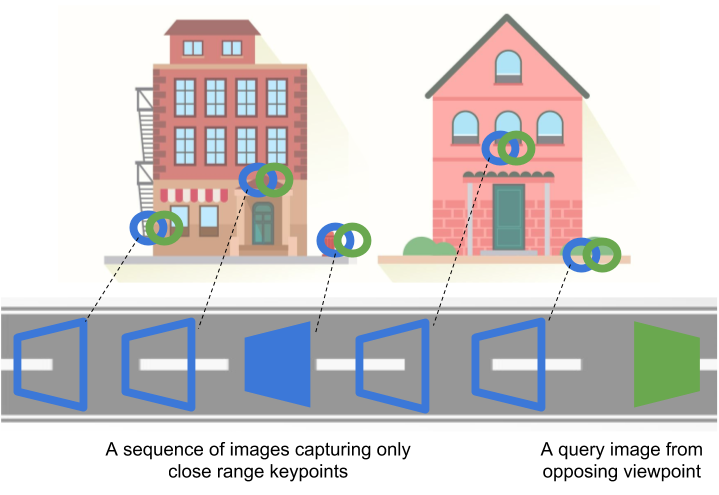}
 \includegraphics[scale=0.29]{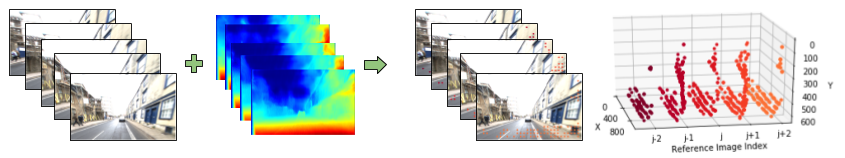}
 \caption{\emph{Top:} Our proposed method uses a \emph{sequence-to-single} image matching technique to accrue depth-filtered keypoints over a sequence of reference frames (blue) to match with a single query image (green) from opposing viewpoint. \emph{Bottom:} The reference frame sequence is combined with the depth masks to obtain a local topo-metric representation.}
 \label{fig:frontFig}
\end{figure}

The color-only VPR techniques for varying environmental conditions and viewpoints have shown great promise~\cite{garg2018lost,gawel2017x,naseer2017semantics,arandjelovic2016netvlad}. However, the performance achieved using these systems might be limited due to the use of only RGB information. Additional information, especially depth, is an obvious potential source for improving performance. Moreover, depth is potentially useful for scale-aware metric pose estimation~\cite{dharmasiri2018eng} though it is not the focus of this work. The use of single-view depth enables a good compromise between image space only methods~\cite{gawel2017x,garg2018lost,arandjelovic2016netvlad} and those based on full 3D reconstruction~\cite{mur2017orb,engel2014lsd}, where it can be challenging for the latter to deal with non-uniform appearance variations~\cite{alismail2016direct} and noisy depth measurements. In the same vein, our proposed system utilizes depth information within a sequence of images to create an \emph{on-demand local topo-metric} representation~\cite{badino2011visual,bazeille2011incremental} whilst not being affected by noise in depth estimation as shown in Figure~\ref{fig:frontFig}.

Our key contributions include:
\begin{itemize}
    \item a novel \emph{sequence-to-single} image matching technique\footnote{The source code will be made publicly available.} that exploits within-the-image information accrued over time to maximize visual overlap between images taken from opposite viewpoints,
    \item a depth-based keypoint filtering technique to use close-range keypoints for dealing with perceptual aliasing under extreme appearance variations,
    \item performance characterization with respect to depth-based keypoint filtering, reference frame sequence length, degree of appearance change, camera motion and depth noise sensitivity.
\end{itemize}

The existing methods for visual place recognition that use a global image descriptor, such as those learnt using deep metric learning techniques~\cite{leepointnetvlad,arandjelovic2016netvlad} encode spatial and structural features implicitly. While these methods form a strong baseline for top candidates retrieval for a query image, the explicit use of geometry can be beneficial for validating these top matching candidates~\cite{garg2018lost,garg2018don't}.

The problem of opposite viewpoints is often dealt with using sensor hardware solutions such as LiDARs~\cite{wolcott2017robust} and panoramic cameras~\cite{arroyo2014bidirectional}. However, humans solve this problem regularly with a limited field-of-view vision and without having to constantly turn around. Under a similar constraint of limited field-of-view vision, dealing with 180 degree viewpoint change along with extreme appearance variations becomes a challenging research problem.

To address this challenge, we propose a \emph{sequence-to-single} matching approach where a sequence of frames centered at the matching candidate location is considered for a single query image. Each of the frames in the reference candidate sequence only contributes keypoints which are within a certain depth range. As we are considering the place recognition problem for opposing viewpoints, it can be assumed that amount of visual overlap between any two matching places will only be limited to a certain depth range (on an average $30-40$ meters for maximum visual overlap~\cite{garg2018lost}). Therefore, the sequence of these depth-filtered keypoints tends to capture visual information that significantly increases the overlap with the query image. We show that the proposed technique consistently outperforms the state-of-the-art methods and also improves performance over the vanilla scenario where a single reference image is used for matching. However, the performance trends depend on the environmental conditions of the compared traverses, where extreme appearance variations tend to benefit the most due to high perceptual aliasing.


\section{Literature Review}

In visual place recognition, changing environments are a major challenge due to perceptual aliasing between the representations of places under severe appearance variations. A number of techniques have been proposed to improve the localization ability, including: leveraging temporal information \cite{milford2012seqslam,liu2013towards,hansen2014visual}, learning the appearance change over time \cite{neubert2013appearance,Lowry2014Morn2Arvo,chen2017deep}, and extracting geometric and spatial information out of an image \cite{Klein2007PTAM,mur2015orb,2017DeMoN}. 

In SeqSLAM \cite{milford2012seqslam}, utilizing sequences of recent images enables localization in challenging environments, by leveraging the scene similarity across recent images. Other sequence approaches include SMART \cite{pepperell2014all}, using network flows \cite{naseer2018robust}, and applying Conditional Random Field theory \cite{cadena2012robust}. These approaches use a variety of image processing front-ends, ranging from Sum of Absolute Differences to more sophisticated methods like HOG \cite{dalal2005histograms} and Convolutional Neural Networks (CNNs)~\cite{krizhevsky2012imagenet}.

CNNs have demonstrated both appearance and viewpoint robustness when applied to visual place recognition~\cite{sunderhauf2015place}. In addition to using off-the-shelf CNNs \cite{sunderhauf2015performance}, the place recognition task has also been learnt end-to-end \cite{arandjelovic2016netvlad}. In HybridNet, an off-the-shelf CNN is improved by training on appearance-change images \cite{arandjelovic2016netvlad,chen2017deep}. In HybridNet~\cite{chen2017deep}, pyramid pooling is used to improve the viewpoint robustness, however this pooling method is still defined by specific regions within a feature map. As an improvement, deep-learnt features have been intelligently extracted out of keypoint locations within these feature maps \cite{ufer2017deep}. \cite{liu2015treasure} proposes cross-convolutional pooling, by pooling features using the spatial position of activations in the subsequent layer. In \cite{chen2017only}, keypoint locations are extracted out of a late convolutional layer, whereas LoST \cite{garg2018lost} uses semantic information to represent places and obtain keypoint locations. LoST improves the localization ability across viewpoint variations as extreme as front to rear-view imagery. However, it relies mostly on visual semantics and spatial layout verification to achieve high performance; in this work, we delve deeper into the efficacy of CNN-based keypoints and descriptors for VPR under challenging scenarios by using depth- and temporal-aware system. 

The use of geometric information for VPR has been shown to improve performance in recent works~\cite{gawel2017x,garg2018lost}. In the same vein, we explore the use of depth information for improving VPR performance under vast variations in appearance and viewpoint. Depth estimation is best achieved using stereoscopic images \cite{2014DepthStereo}, however recent advances have enabled depth estimation from monocular images. Geometric constraints \cite{Reza2018UnsuperDepth} and non-parametric sampling \cite{Karsch2014NonParaDepth} have been used to extract depth out of a monocular image. Improved performance can be gained by training a CNN to estimate pixel-by-pixel depth within an image \cite{Lui2016LearningDepth}. In recent years, several deep networks have been proposed to estimate per-pixel-depth map and visual odometry. For instance,\cite{Eigen2014Depth} is one of the earliest works in predicting depth by regressing with the ground truth for monocular images. The authors in \cite{2017DeMoN} additionally use the motion parallax between image pairs, enabling robust depth estimation for novel scenes. Rather than using supervised learning to estimate depth, \cite{garg2016unsupervised, babu2018deeper} use unsupervised learning in an end-to-end framework. CNN-based depth estimation has been employed for dense monocular slam and semantic 3D reconstruction~\cite{Keisuke2017cnnslam}. Our proposed approach exploits depth estimates to filter keypoints from a sequence of reference frames that are beyond a certain range for the expected visual overlap between front- and rear-view images. This forms a local topo-metric representation of the reference candidate match location which is a good compromise between full 3D reconstruction and single image only methods. However, methods like~\cite{Keisuke2017cnnslam} will be complementary to our proposed approach.

\section{Proposed Approach}
We use a hierarchical place matching pipeline~\cite{garg2018lost} where top-N candidate matches are generated using cosine distance based descriptor comparison of a query image with the reference database using state-of-the-art retrieval methods~\cite{arandjelovic2016netvlad,garg2018lost,babenko2015aggregating,liu2015treasure}. The query image is then matched with these top-N candidates using depth-filtered keypoints extracted from a sequence of reference frames centered at the candidate index. Figure~\ref{fig:flowchart} shows a flow diagram of our proposed approach as explained in subsequent sections.

\subsection{Keypoint and Descriptor Extraction}
For a given pair of matching images, we extract keypoints and descriptors from the \emph{conv5} layer of the ResNet101~\cite{lin2016refinenet} as described in details in~\cite{garg2018lost} and briefly here for sake of completion. For a \emph{conv5} tensor of dimension $W\times H\times C$, representing width, height, and the number of channels (or feature maps) of the tensor, a keypoint location is determined for each of the channels based on the maximally-activated 2D location within that channel. Using the same tensor, a descriptor at the keypoint location is extracted as a $C$ dimensional vector along the third axis. As only one keypoint is extracted per channel, we obtain a total of $C$ such keypoints and descriptors. For a given pair of images, the correspondences between the keypoints are assumed based on their channel index.

\subsection{UNsupervised MOnocular Depth Estimation}
Given the reference database images, a per-pixel-depth map is estimated using the framework described in UnDEMoN~\cite{babu2018deeper}. The network estimates disparity maps which are then converted into depth maps using the camera parameters of the training dataset.
For a keypoint $k$ at a pixel location $p_i^k$, depth $u_{p_i^k}$ is estimated from the disparity $u'_{p_i^k}$ using the formula $u_{p_i^k} = \frac{fb}{u'_{p_i^k}}$ where $f$ is the focal length and $b$ is the baseline distance between the stereo image pairs from the KITTI dataset~\cite{Geiger2012kitti} used to train the model. In this work, we directly used the KITTI trained models from~\cite{babu2018deeper} without any fine tuning performed on the datasets used in our experiments.

\begin{figure}
    \centering
    \includegraphics[scale=0.28]{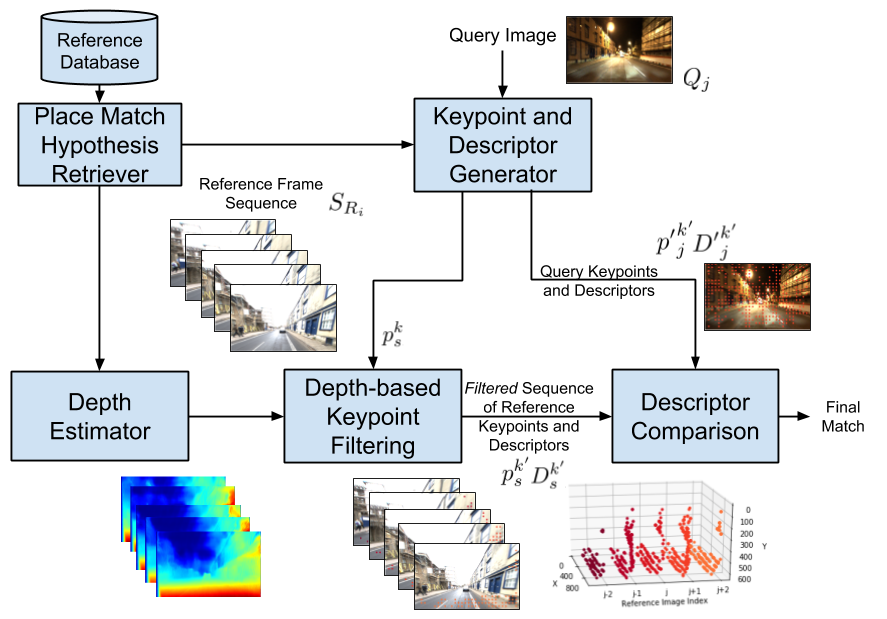}
    \caption{Flowchart of our proposed approach.}
    \label{fig:flowchart}
\end{figure}

\subsection{Depth-Filtered Keypoints Sequence}
For a query image, $Q_j$ and a top matching reference candidate $R_i$, keypoints ${p'}_j^k$ and $p_i^k$ and descriptors ${D'}_j^k$ and $D_i^k$ are extracted respectively. Further, a sequence of reference frames $S_{R_i}$ of length $l$ is considered across $R_i$, that is, $S_{R_i}=[R_{i-l/2},R_{i+l/2}]$ as shown in Figure~\ref{fig:flowchart}. Then, a depth range threshold $d$ is used to filter out the keypoints that are far from the camera. Therefore, a set of keypoints is obtained from the reference sequence as below:

\begin{equation}
 p_s^{k'} = \{p_s^k \mid u_{p_s^k} < d\} \quad \forall \hspace{0.2cm}  k \in C, s \in S_{R_i} 
\end{equation}
 
where $k'$ is a filtered keypoint index within $C'\subseteq C$. 

\subsection{Descriptor comparison}
For each of these keypoints $p_s^{k'}$ in the reference sequence, there exists a corresponding keypoint in the query image\footnote{The correspondences are based on the channel (feature map) index ($k'$) within the \emph{conv5} tensor}. We use cosine distance between the descriptors of these corresponding keypoints to obtain minimum descriptor distance from the query keypoint to the corresponding keypoint in any of the reference sequence frames:

\begin{equation}
  r_{ji}^{k'} =  \min_s( 1 - \frac{{D'}_j^{k'} \cdot D_s^{k'}   }{\lVert {D'}_j^{k'}  \rVert_2 \lVert D_s^{k'}  \rVert_2}) \quad \forall \hspace{0.2cm} k' \in C', s \in S_{R_i}
\end{equation}

An average over all the filtered keypoints' descriptor distances $r_{ji}^{k'}$ gives the matching score $r_{ji}$ between the query image and the reference candidate. 

The least scoring candidate is then considered as the final image match for the query.

\section{Experimental Setup}
\subsection{Oxford Robotcar Dataset}
The Oxford Robotcar Dataset~\cite{maddern20171} comprises traverses of Oxford city during different seasons, time of day and weather conditions, capturing images using cameras pointing in all four directions. We used an initial 2.5 km traverse from front- and rear-view cameras for four different environmental settings: Overcast Autumn, Night Autumn, Dawn Winter and Overcast Summer\footnote{2014-12-09-13-21-02, 2014-12-10-18-10-50, 2015-02-03-08-45-10 and 2015-05-19-14-06-38 respectively in~\cite{maddern20171}}. We used the GPS data to sample image frames at a constant distance of approximately $2$ meters.

\begin{figure}[t]
\vspace{0.2cm}
\centering
\begin{tabular}{c}
 \includegraphics[scale=0.17]{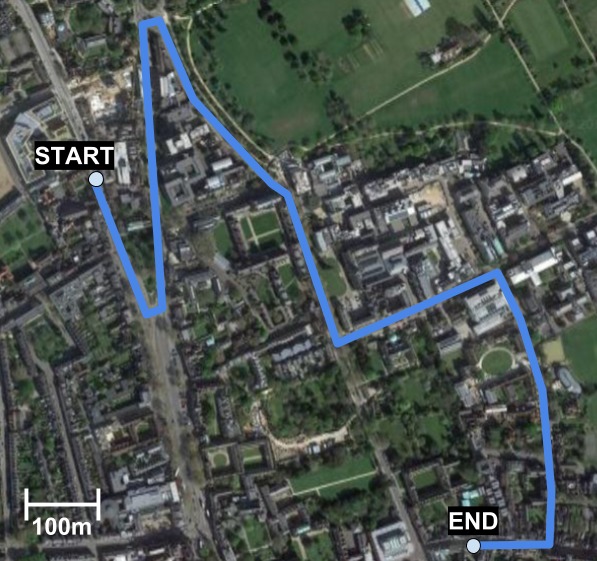}
\end{tabular}
\caption{Aerial view of ground truth trajectories for Oxford Robotcar dataset. Source: Google Map}
\label{fig:gt_aerial_trajs}
\end{figure}

\newcommand{\scaleOne}{0.23}
\begin{figure*}
\begin{tabular*}{\textwidth}{cccc}
 \includegraphics[scale=\scaleOne]{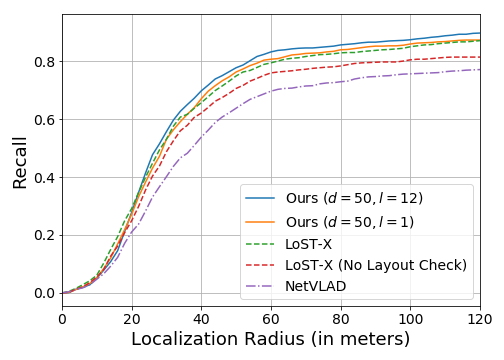} &
 \includegraphics[scale=\scaleOne]{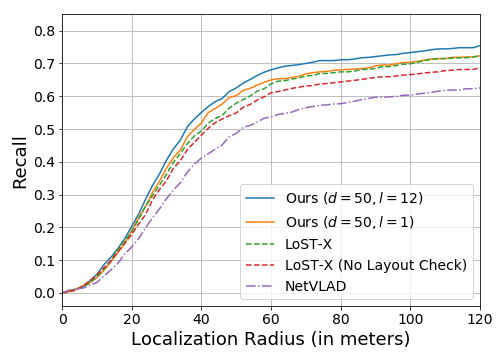}&
 \includegraphics[scale=\scaleOne]{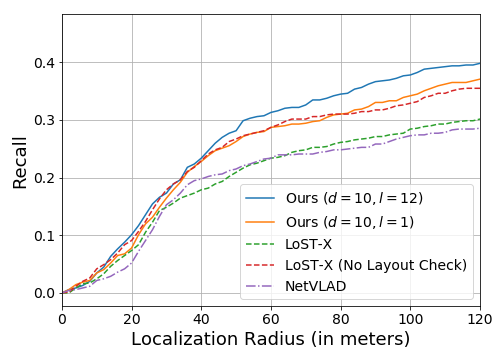} &
 \includegraphics[scale=\scaleOne]{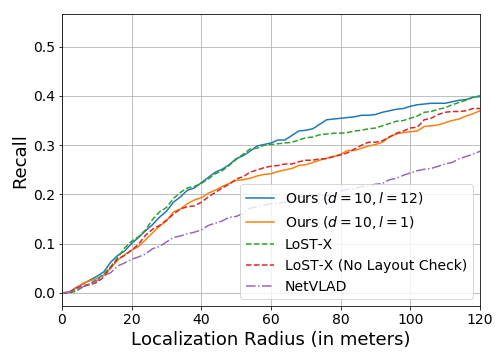} \\
 (a) Overcast Autumn & (b) Overcast Summer & (c) Dawn Winter & (d) Night Autumn  \\
\end{tabular*}
\caption{Performance comparison against state-of-the-art methods for opposite viewpoints and changing conditions. These front-view imagery from the four traverses: (a) Overcast Autumn, (b) Overcast Summer, (c) Dawn Winter, and (d) Night Autumn was compared against the rear-view imagery from Overcast Autumn traverse.}
\label{fig:compareSota}
\end{figure*}

\newcommand{\scaleTwo}{0.06}
\newcommand{\scaleTwoHalf}{0.12}
\begin{figure*}
 \begin{tabular*}{\textwidth}{c|cccccc}
 
  &
  \includegraphics[scale=\scaleTwoHalf]{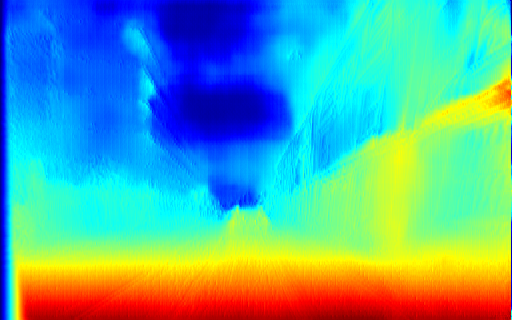} &
  \includegraphics[scale=\scaleTwoHalf]{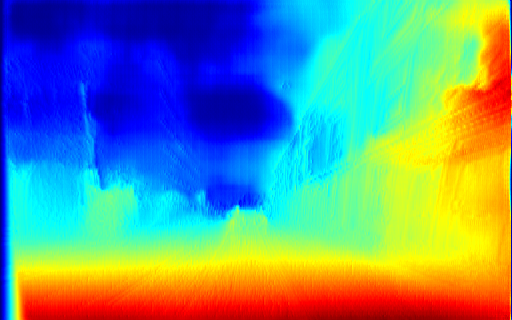} &
  \includegraphics[scale=\scaleTwoHalf]{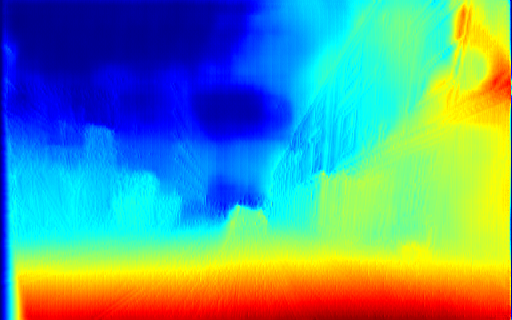} &
  \includegraphics[scale=\scaleTwoHalf]{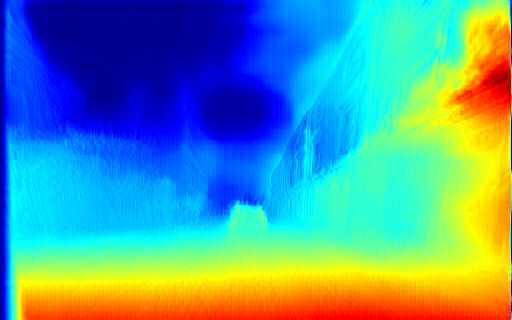} &
  \includegraphics[scale=\scaleTwoHalf]{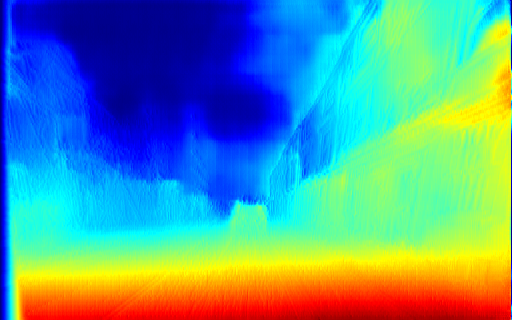} & \\
  
  \includegraphics[scale=\scaleTwo]{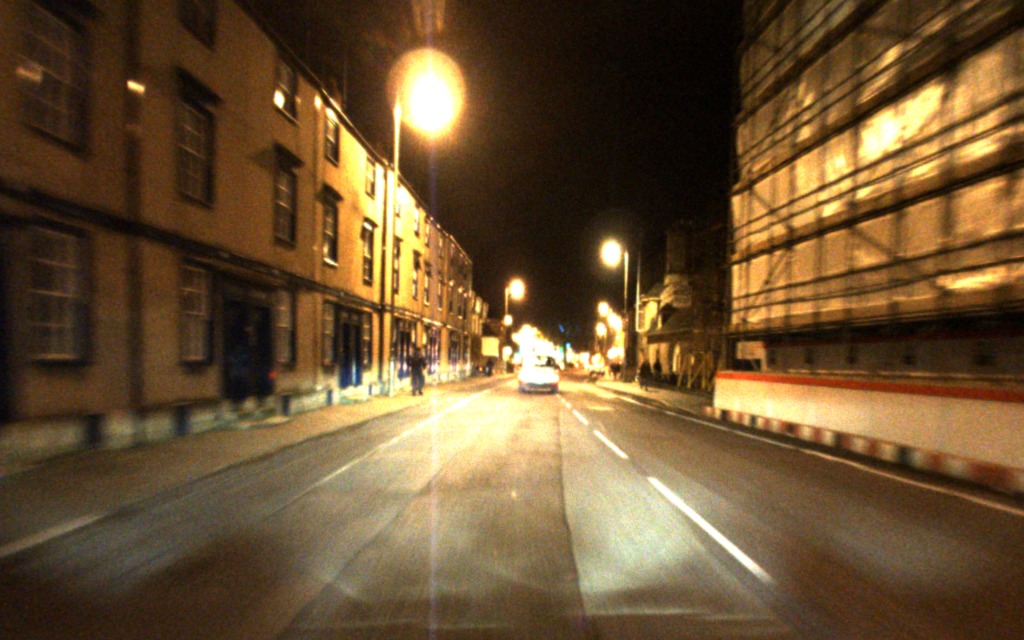} &
  \includegraphics[scale=\scaleTwo]{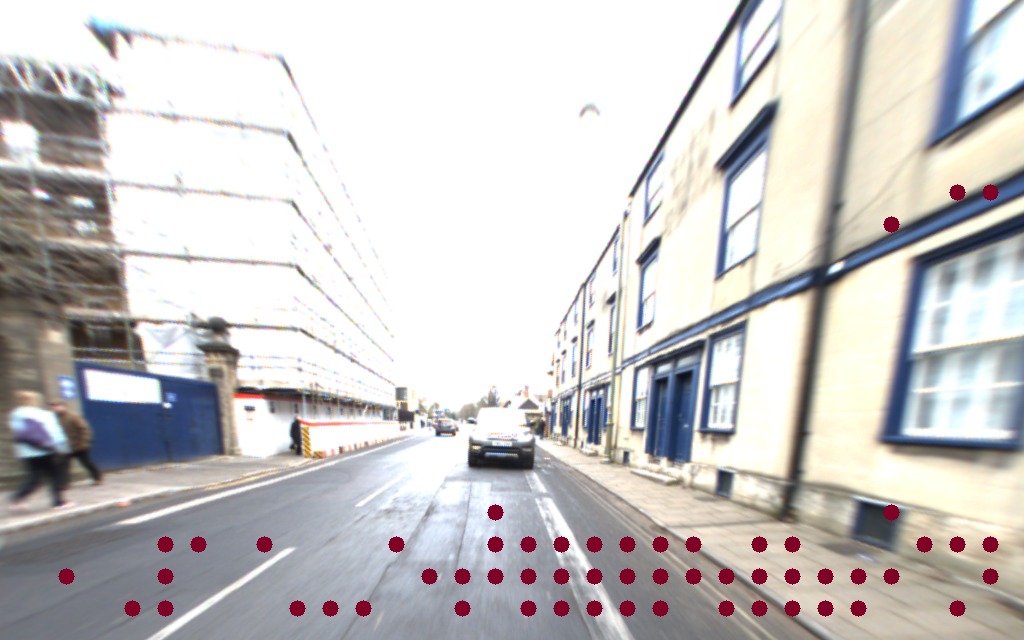} &
  \includegraphics[scale=\scaleTwo]{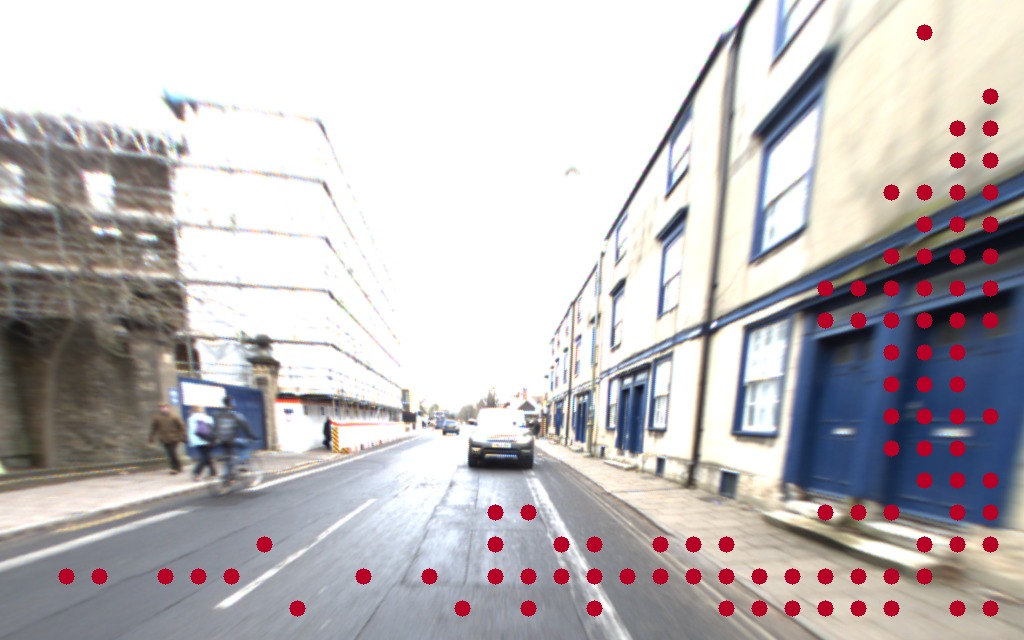} &
  \includegraphics[scale=\scaleTwo]{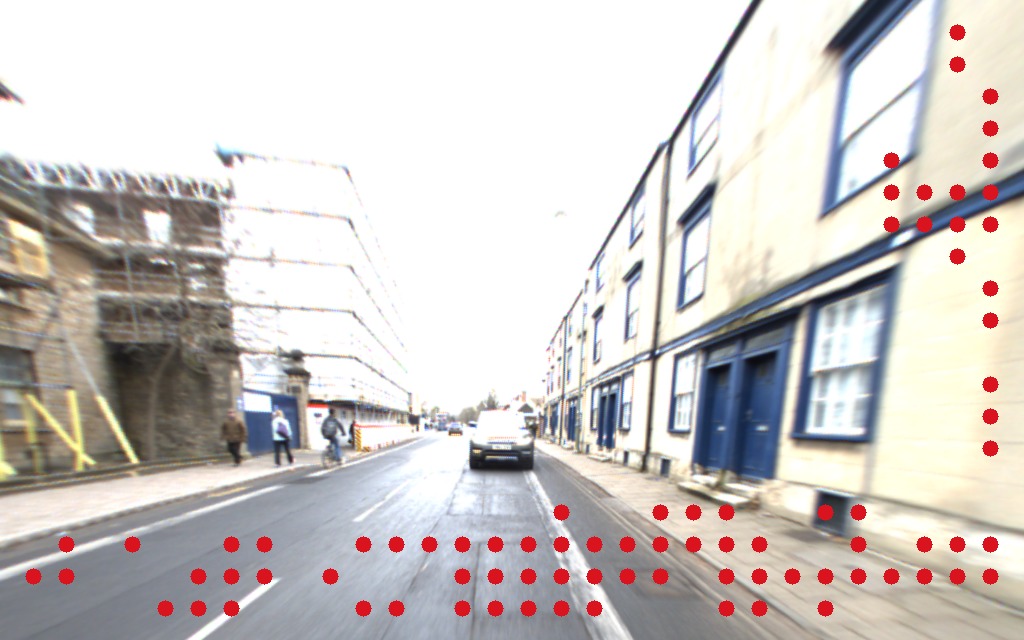} &
  \includegraphics[scale=\scaleTwo]{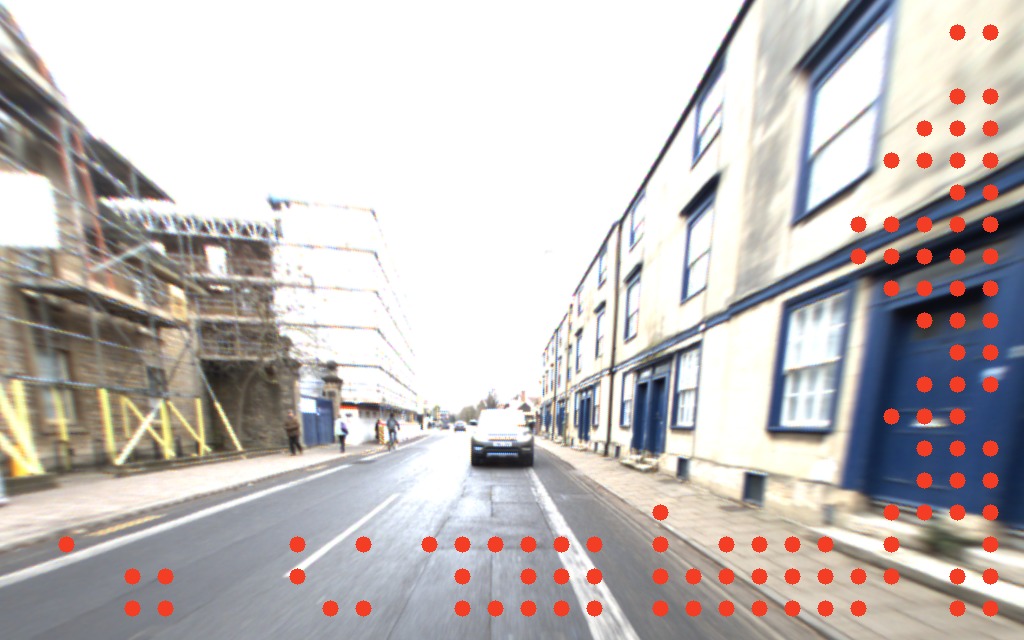} &
  \includegraphics[scale=\scaleTwo]{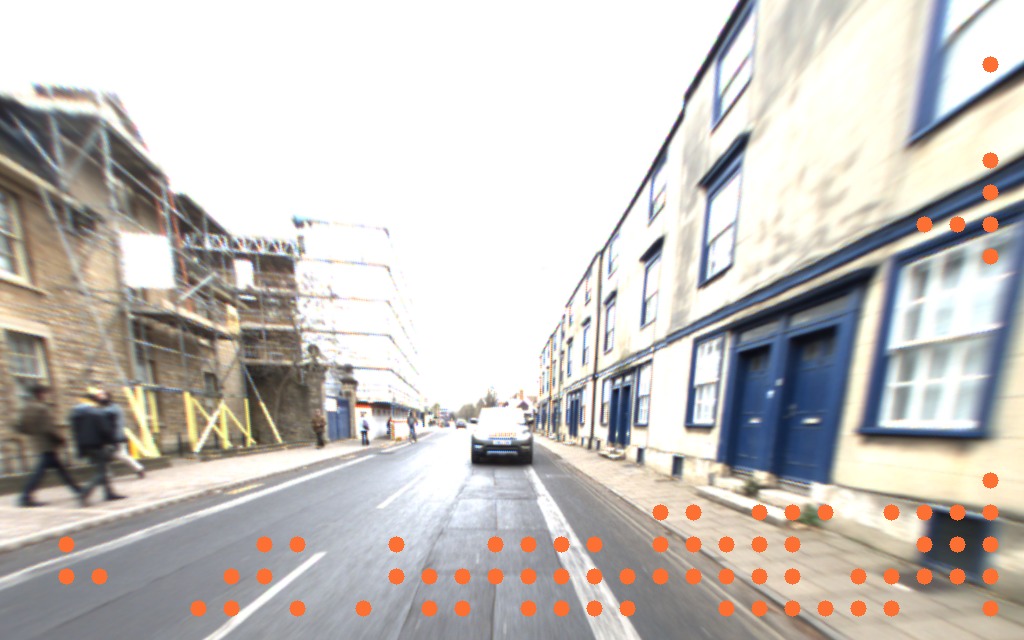} &
  \includegraphics[clip, trim=2cm 0cm 0cm 2cm,scale=0.25]{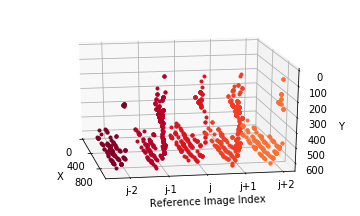}\\
  
  \includegraphics[scale=\scaleTwo]{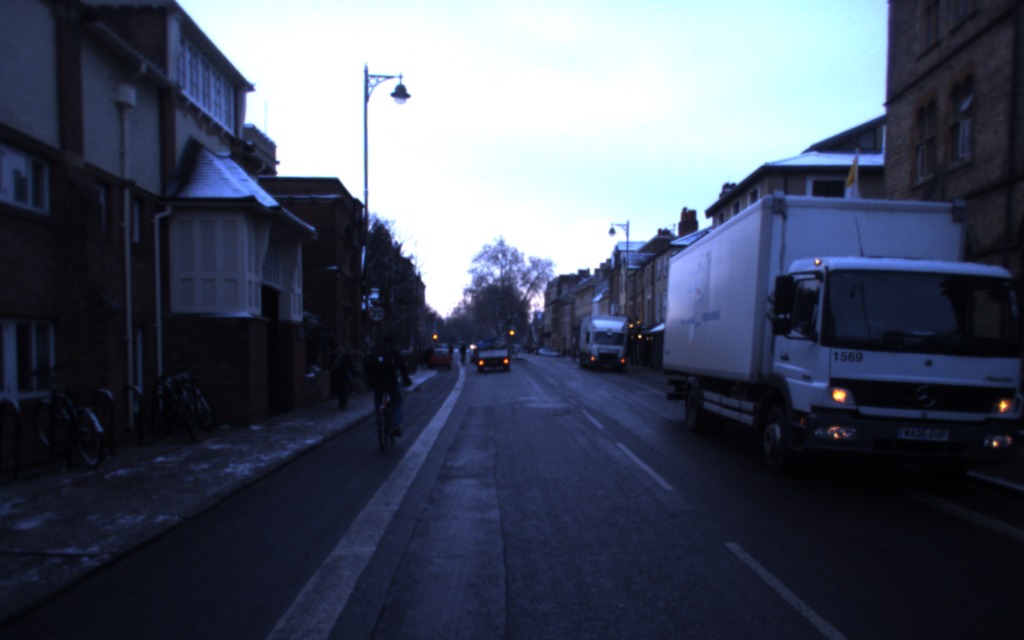} &
  \includegraphics[scale=\scaleTwo]{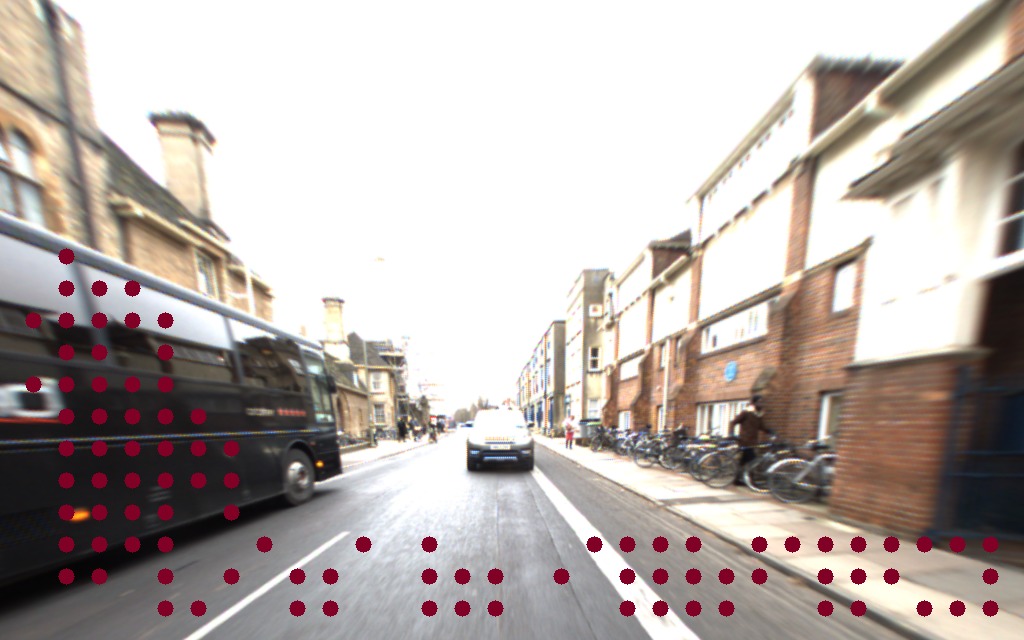} &
  \includegraphics[scale=\scaleTwo]{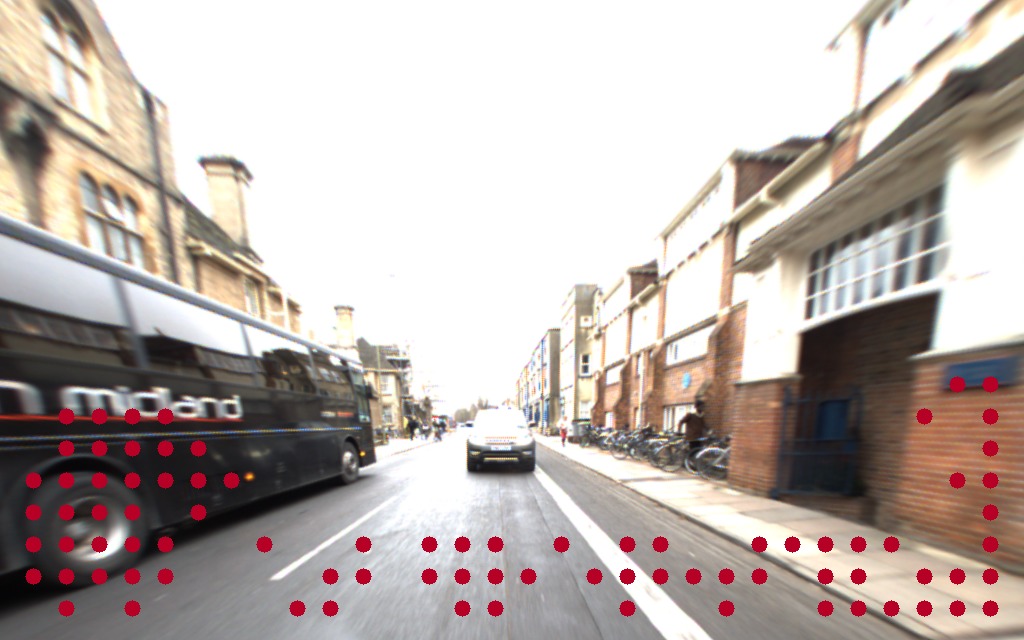} &
  \includegraphics[scale=\scaleTwo]{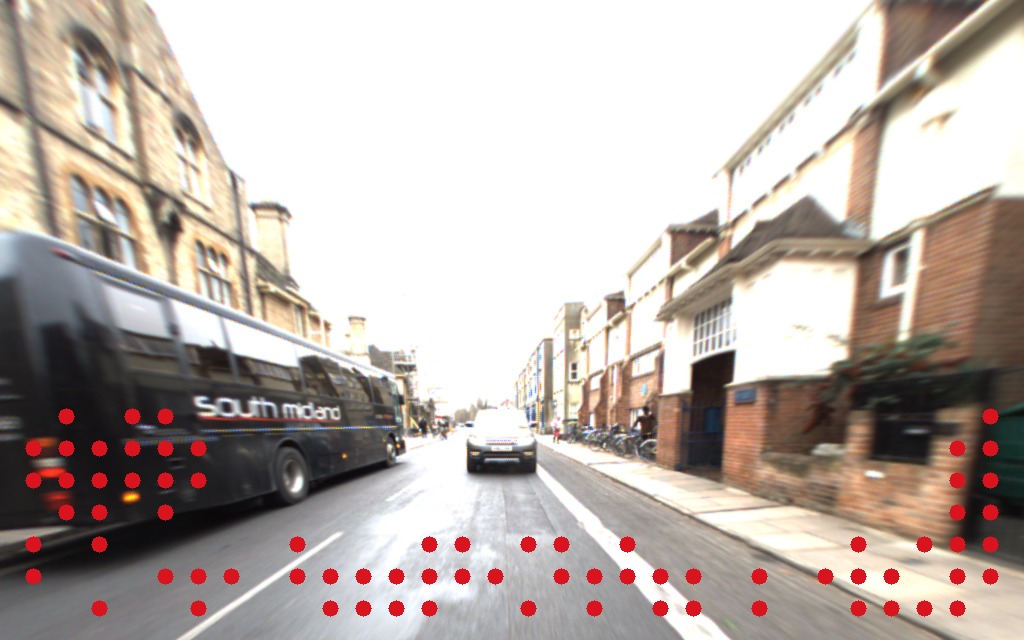} &
  \includegraphics[scale=\scaleTwo]{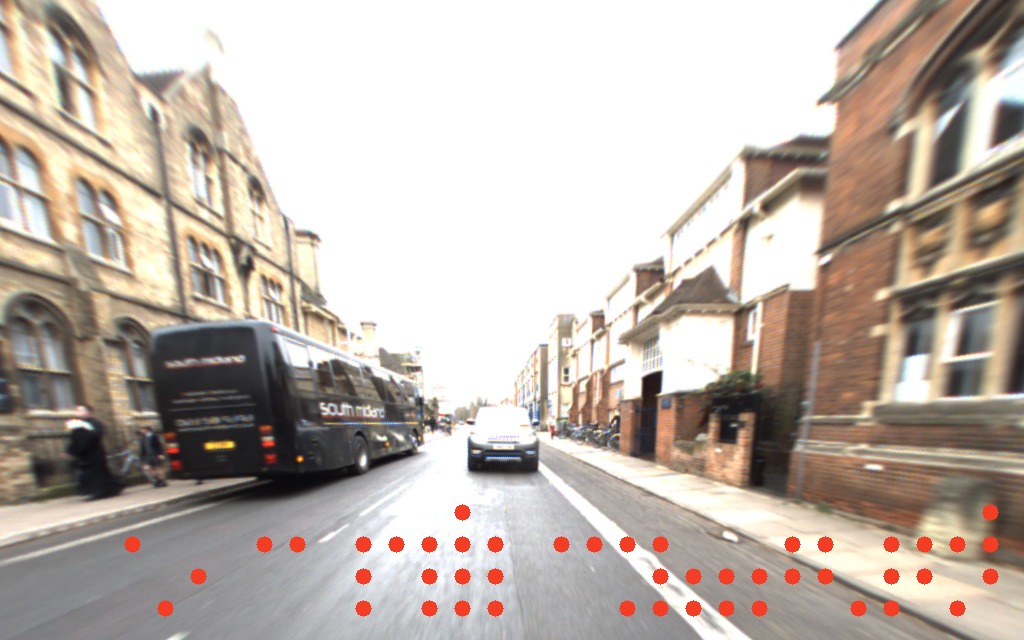} &
  \includegraphics[scale=\scaleTwo]{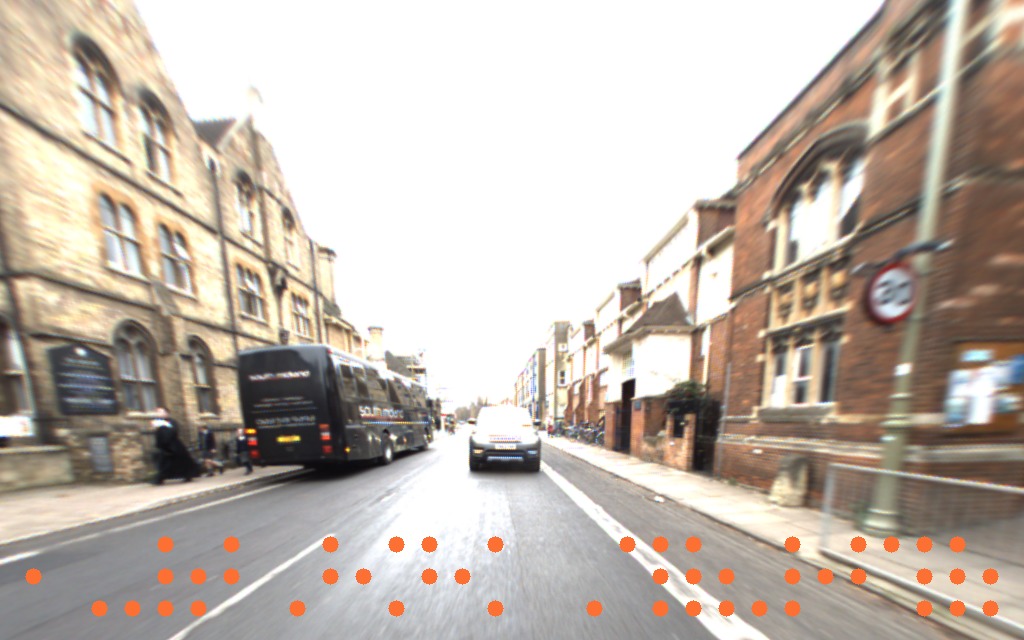} &
  \includegraphics[clip, trim=2cm 0cm 0cm 2cm,scale=0.25]{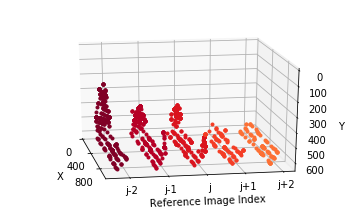}\\
  

  \includegraphics[scale=\scaleTwo]{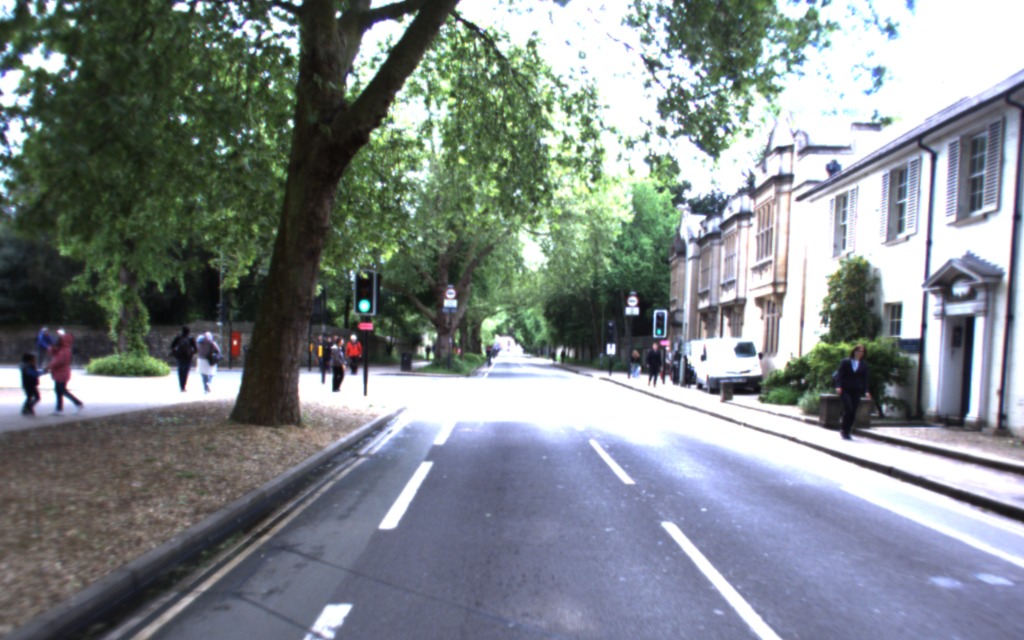} &
  \includegraphics[scale=\scaleTwo]{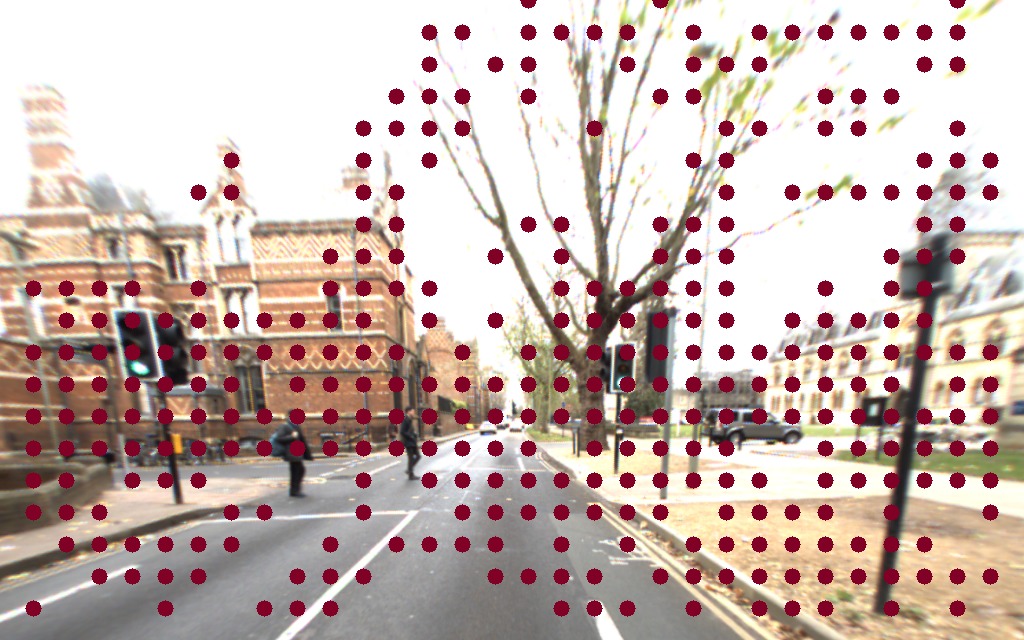} &
  \includegraphics[scale=\scaleTwo]{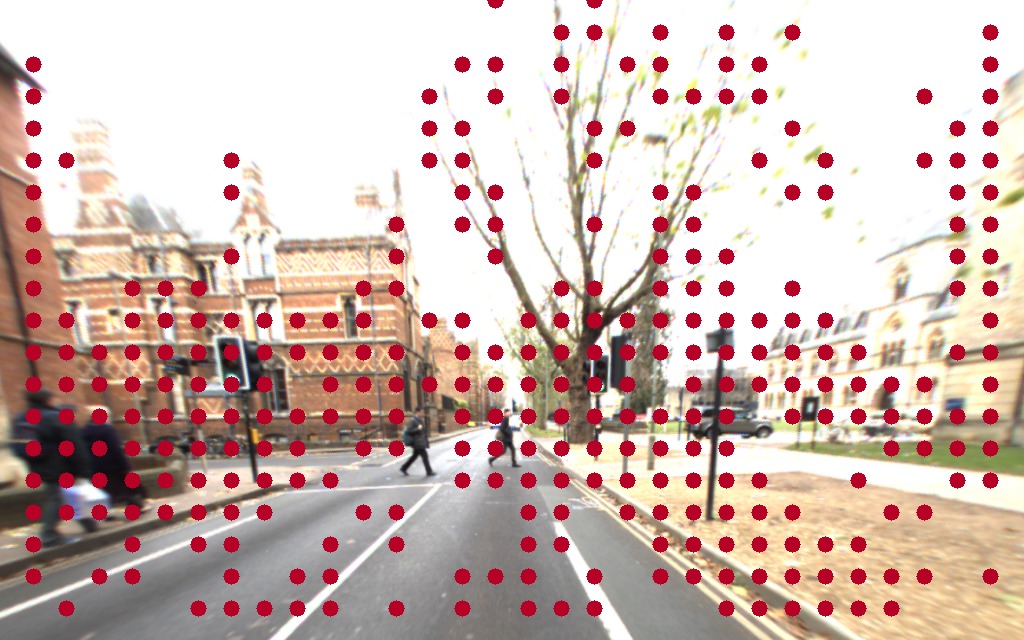} &
  \includegraphics[scale=\scaleTwo]{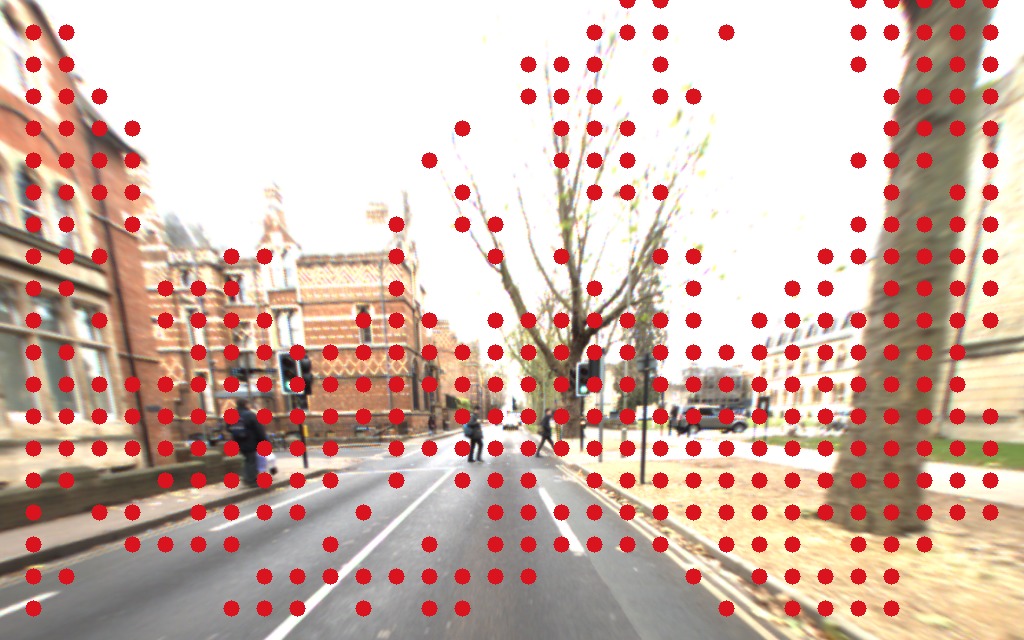} &
  \includegraphics[scale=\scaleTwo]{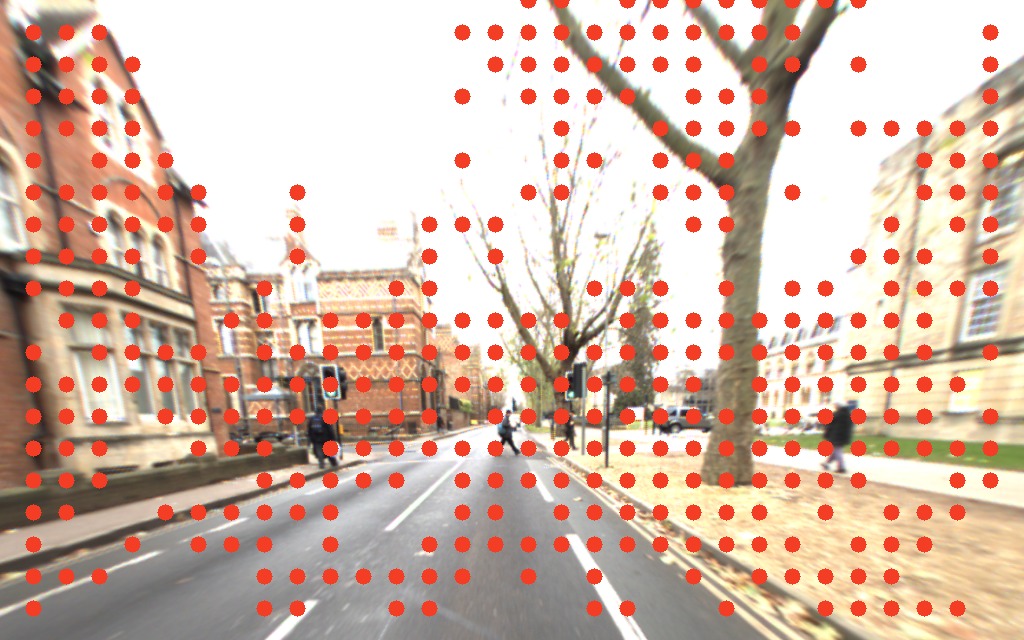} &
  \includegraphics[scale=\scaleTwo]{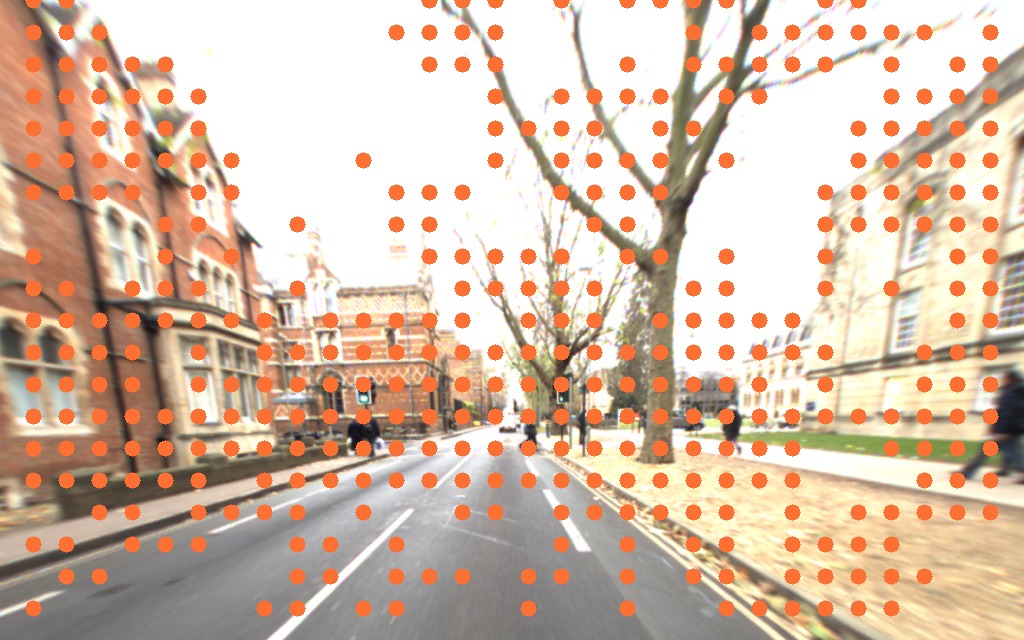} &
  \includegraphics[clip, trim=2cm 0cm 0cm 2cm,scale=0.25]{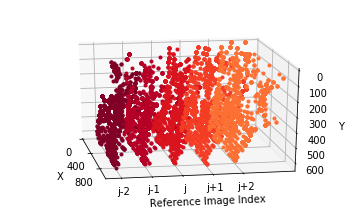}\\
  
  $i$ &  $j-2$ & $j-1$ & $j$ & $j+1$ & $j+2$ \\
  
  Query Index & \multicolumn{5}{c}{Reference Frame sequence centered at $j$} \\
  
  \end{tabular*}
 \caption{Example Matches: Keypoints from the front-view query images (\emph{leftmost column}) are matched against a reference frame sequence using only depth-filtered keypoints (marked in different colors) accrued over the sequence. The \emph{rightmost column} shows the sequential collection of these keypoints such that the horizontal axis topologically connects the sequence of images, however, for each image metric depth is used to separate the points. The \emph{topmost row} shows depth masks for the reference image sequence used in the second row.}
 \label{fig:QualPerf}
\end{figure*}

\subsection{Ground Truth and Evaluation Method}
We use GPS information to generate place recognition ground truth for the Oxford traverses. However, unlike the forward-forward image matching scenario, the opposite viewpoint place matching involves a \emph{visual offset} of approximately $30-40$ meters where the visual overlap between the matching pair of images becomes maximum, as described in \cite{garg2018lost}. We use recall as a performance measure, that is equal to the number of true positives detected divided by the total number of positives in the dataset comparison. A match is considered a true positive if it lies within a certain radius of its ground truth GPS location and this radius is varied to generate the performance curves. 

\subsection{Comparative Study}
We compare our proposed approach with two state-of-the-art methods: `NetVLAD'~\cite{arandjelovic2016netvlad} and `LoST-X'~\cite{garg2018lost}. LoST-X uses spatial layout matching of semantically-filtered keypoints; we also include the `LoST-X (No Layout Check)' version in order to compare against semantic filtering based \emph{conv5} descriptor matching. For our proposed approach, we show results for two scenarios: one with the reference frame sequence length $l$ set to $1$ which means traditional single-to-single image matching, and the other with $l$ set to $12$ which is approximately $25$ meters for the Oxford dataset. The depth range threshold is set to $50$ and $10$ meters for similar and different environmental conditions respectively. The choice of parameters is primarily based on the previous study~\cite{garg2018lost} that shows $30-40$ meters to be an average distance between two images from opposing viewpoints leading to the maximum visual overlap. Further, in Section~\ref{sec:Res_PerfChar}, we show performance characteristics of our proposed system with respect to these parameters that support choosing depth threshold depending on the degree of appearance variations. For a fair comparison, for all the methods, we use top $5$ match hypothesis generated by cosine distance comparison of NetVLAD descriptors.

\section{Results}

\subsection{Performance comparison}
Figure~\ref{fig:compareSota} shows performance comparison of our proposed approach against state-of-the-art methods for opposite viewpoint image matching under varying appearance of the environment. It can be observed that our proposed method consistently outperforms both NetVLAD and LoST-X. The performance gains are even more when compared against LoST-X without employing its spatial layout verification. This indicates that spatial layout matching is also a key step to attain higher performance and can potentially complement our proposed approach to improve performance even further. The performance curves also show consistently that using only a single reference image based matching has sub-optimal performance than the reference frame sequence based matching.

\subsubsection*{Qualitative Results} Figure~\ref{fig:QualPerf} shows qualitative matches for opposite viewpoint image matching under varying environmental conditions. A sequence of frames centered at the candidate match location (index=$j$) are shown with their keypoints within a certain depth range. The \emph{rightmost column} shows the sequential collection of these keypoints such that the horizontal axis topologically connects the sequence of images, however, for each image metric depth is used to separate the points. The \emph{topmost row} shows depth masks for the reference image sequence used in the second row. It can be observed that due to imperfections in the estimated depth, filtering of keypoints within a given depth range is not always consistent, however, due to visual overlap between consecutive frames, this doesn't pose a problem.

\begin{figure}
 \begin{tabular}{cc}
  \includegraphics[clip, trim=5.5cm 2cm 2cm 3cm,scale=0.27]{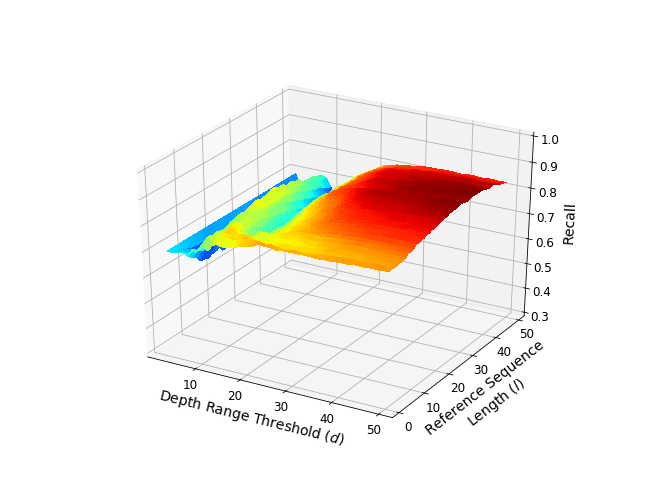}&
  \includegraphics[clip, trim=5.5cm 2cm 2cm 3cm,scale=0.27]{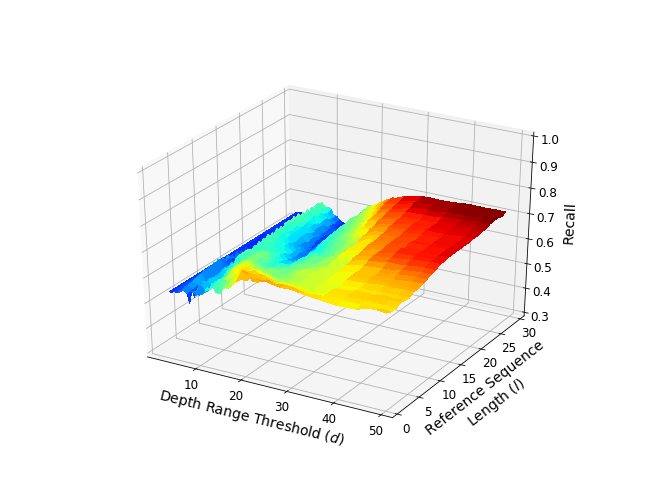}
 \end{tabular}
\caption{\emph{Similar environmental conditions}: The performance curves comparing Autumn Day Rear with Autumn Day Front (\emph{left}) and (b) Summer Day Front (\emph{right}) show that performance tends to increase as more visual information becomes available by considering distant points from the camera using depth information and a longer sequence of reference frames.}
\label{fig:3d-sameCond}
\end{figure}

\begin{figure}
 \begin{tabular}{cc}
  \includegraphics[clip, trim=5.5cm 2cm 2cm 3cm,scale=0.27]{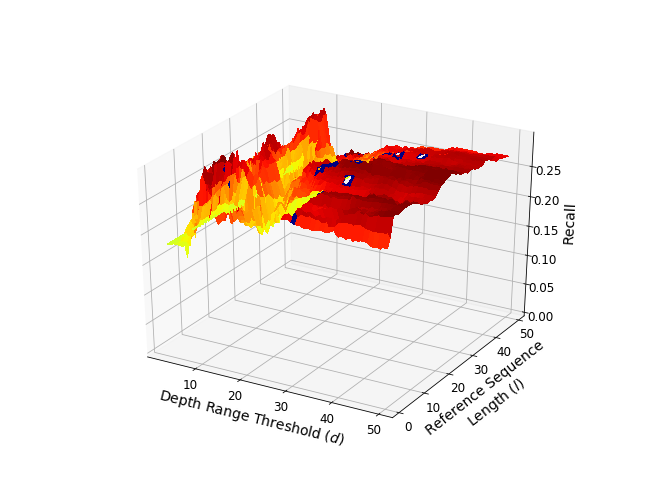}&
  \includegraphics[clip, trim=5.5cm 2cm 2cm 3cm,scale=0.27]{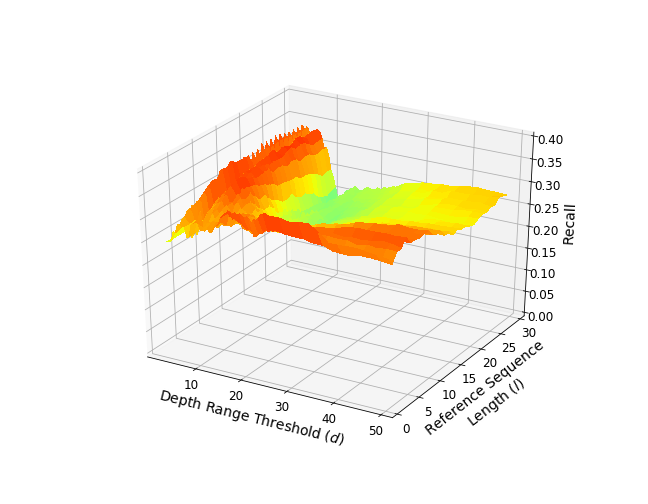}
 \end{tabular}
\caption{\emph{Different environmental conditions}: The performance curves comparing Autumn Day Rear with Night Front (\emph{left}) and Winter Front (\emph{right}) show that distant points from the camera tend to decrease the performance. Further, longer reference frame sequence does not provide any significant performance gains unlike those observed when environmental conditions remained similar as shown in Figure~\ref{fig:3d-sameCond}.}
\label{fig:3d-oppCond}
\end{figure}

\subsection{Performance Characteristics}
\label{sec:Res_PerfChar}
In this section, we show the performance characteristics of our proposed system with respect to the two system parameters: depth range threshold, $d$ and reference frame sequence length, $l$. These characterizations are discussed below with respect to the extent of appearance variations due to varying environmental conditions. Further, we show system's characteristics with respect to different camera speed and the quality of depth estimation.

\subsubsection{Moderate Appearance Variations}
Figure~\ref{fig:3d-sameCond} shows that for opposite viewpoints under moderate or no changes in environmental conditions, more visual information in the form of longer reference sequence length and larger pool of keypoints within each of the reference frames, helps attain high performance. This shows that descriptors used from within the CNN are robust to moderate variations in appearance and are able to discriminate between a large number of false and true keypoint correspondences, therefore, able to utilize additional visual information.

\begin{figure*}
 \begin{tabular*}{\textwidth}{ccc}
  \includegraphics[scale=0.22]{new-dr-nf-s0}&
  \includegraphics[scale=0.22]{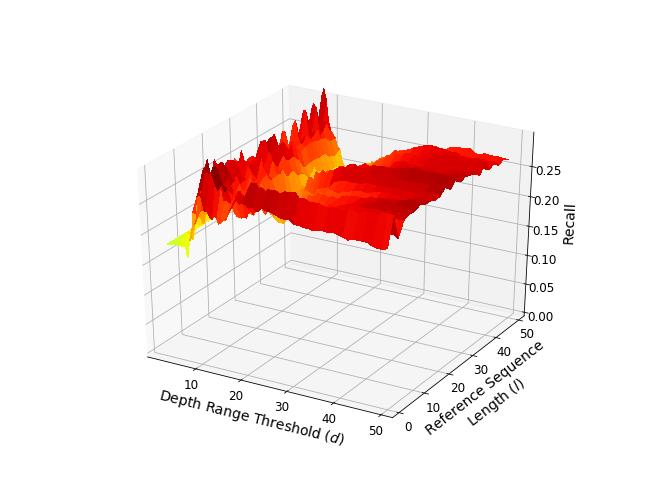}&
  \includegraphics[scale=0.22]{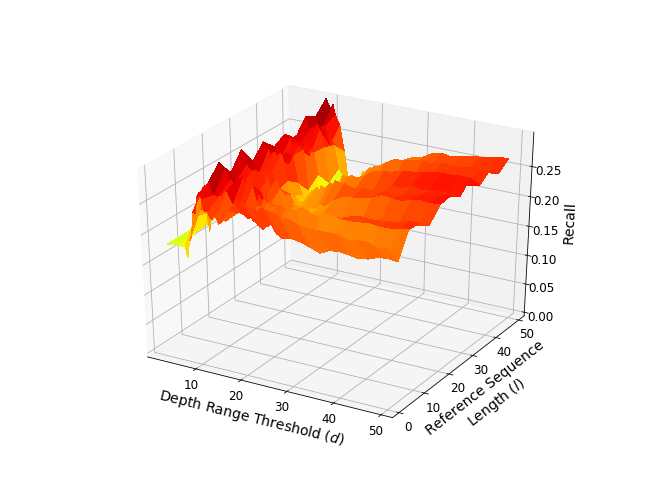} \\
  (a) $1\times$ & (b) $2\times$ & (c) $4\times$ \\
 \end{tabular*}
\caption{\emph{Effects of camera speed}: The performance trends tend to remain similar even when the reference frame sequence is considered at a different camera speed (simulated by skipping frames in the sequence): (a) $1\times$, (b) $2\times$, and (c) $4\times$. It can also be observed that with a higher camera speed (frame skip rate), the peak performance decreases, however, the effect of using a shorter depth range becomes more prominent.}
\label{fig:3d-CamSpeed}
\end{figure*}

\subsubsection{Extreme Appearance Variations}
Figure~\ref{fig:3d-oppCond} shows the performance characteristics for opposite viewpoints under extreme appearance variations from day to night and autumn to winter. It can be observed that, unlike the previous scenario of moderate appearance variations, performance tends to decrease as the visual information tends to increase, especially by allowing more keypoints per reference frame. This can be attributed to high perceptual aliasing caused due to extreme appearance variations. Therefore, accumulating keypoints within a short range of camera over the reference frame sequence helps attain optimal performance. Further within the optimal performance region, the reference frame sequence length tends to have limited effect on the performance, therefore, allowing the selection of a shorter sequence length.

\subsubsection{Camera Speed}
Figure~\ref{fig:3d-CamSpeed} shows variation of performance with respect to changing camera speed for the Day-Front and Night-Rear comparison. This is simulated by skipping frames within the reference frame sequence at: (a) $1\times$, (b) $2\times$, and (c) $4\times$. It can be observed that with a higher camera speed (or frame skip rate), the peak performance decreases, however, the effect of using a shorter depth range becomes more prominent.

\subsubsection{Quality of Depth Estimation}
\begin{figure}
\centering
\begin{tabular}{cc}
    \includegraphics[clip, trim=5.5cm 2cm 2cm 3cm,scale=0.27]{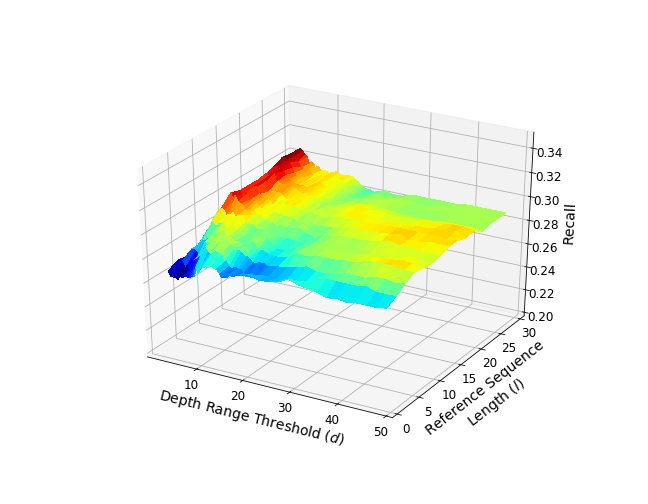}&
    \includegraphics[clip, trim=5.5cm 2cm 2cm 3cm,scale=0.27]{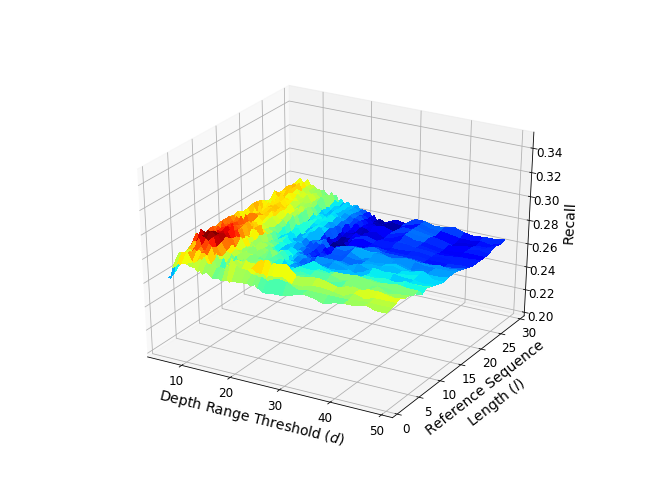}\\
    (a) Ground Truth Depth & (b) Estimated Depth \\
    \vspace{0.1cm}
\end{tabular}
\begin{tabular}{ccc}
\includegraphics[scale=0.13]{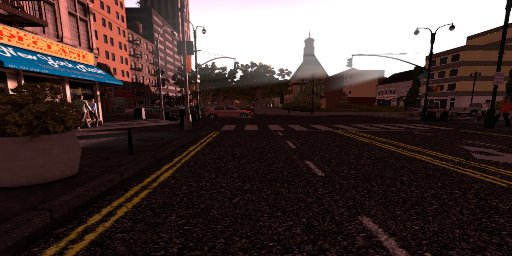} &
\includegraphics[scale=0.13]{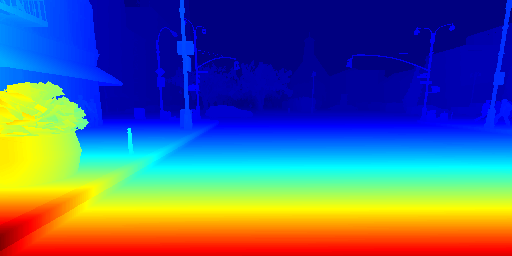} &
\includegraphics[scale=0.13]{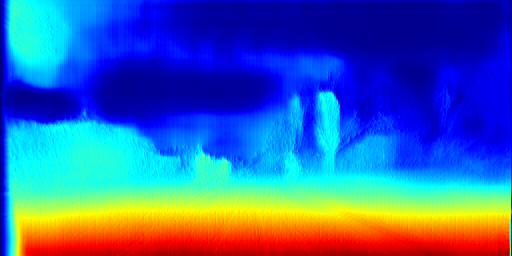} \\
Input Image &  Ground truth Depth &  Estimated Depth 
\end{tabular}
    \caption{Quality of Depth: \emph{Top:} Performance trends tend to remain similar for both (a) Ground truth depth and (b) Estimated depth. However, peak performance is slightly better when using ground truth depth along with a smooth variation with respect to the system parameters. \emph{Bottom:} Ground truth and estimated depth masks for an input image.}
    \label{fig:DepGT}
\end{figure}

We use Synthia dataset~\cite{ros2016synthia} for evaluating the effect of quality of depth\footnote{The depth estimation network used in this paper~\cite{babu2018deeper} is trained on KITTI dataset~\cite{Geiger2012kitti}.} for our proposed approach. For this purpose, we use the front- and rear-view images from the Dawn and Fall traverses of Sequence-02 that have very different environmental conditions, as also utilized in~\cite{garg2018lost}. Figure~\ref{fig:DepGT} shows performance comparison using (a) ground truth depth and (b) estimated depth. It can be observed that the performance trends remain similar to those observed for Oxford dataset for opposite viewpoints and varying appearance. Further, with the use of ground truth depth, peak performance is slightly higher and performance variations are smooth with respect to the system parameters. The bottom row in Figure~\ref{fig:DepGT} also shows a comparison between the ground truth and estimated depth mask for an input image from Synthia reference traverse.

\section{Discussion}
The performance characterizations in Figure~\ref{fig:3d-sameCond} and~\ref{fig:3d-oppCond} show that the use of more visual information by including more distant points and using longer reference frame sequence has varied implications, depending on the variations in the scene appearance. While similar environmental settings tend to benefit from increasing visual information, performance drops drastically when images are matched across different environmental conditions. This drop in performance occurs because of perceptual aliasing among large number of keypoints matched under extreme appearance variations. Our proposed approach allows control of this visual information to achieve optimal performance by using close-range keypoints collected over multiple reference frames.

Most of the existing literature for VPR that deals with challenging appearance variations, generally deals with similar or moderate variations in viewpoints. In such cases, the above finding often gets discounted because the overall structure of the scene remains mostly similar that massively aids in matching, especially when images are down-sampled, for example, patch-normalization in SeqSLAM~\cite{milford2012seqslam}, HoG in \cite{naseer2014robust}, and flattened deep-learnt tensors in \cite{razavian2014cnn,chen2014convolutional,babenko2015aggregating,sunderhauf2015performance}. Furthermore, as we use the deep metric learning method NetVLAD~\cite{arandjelovic2016netvlad} as a baseline to generate top candidate matches, it shows that such methods, despite implicitly encoding the spatial and structural information, still have a large room for improvement. The above analysis helps in gaining insights about the behavior of deep-learnt convolutional descriptors under different appearance conditions when no assumptions are made about the overall structural similarity.

\section{Conclusion and Future Work}
Visual place recognition under opposing viewpoints and varying environmental conditions is a challenging problem and requires effective use of both scene appearance and scene geometry. In this paper, we proposed a \emph{sequence-to-single} matching approach where an on-demand local topometric representation of reference image sequence was used to match with a single query image using depth-based keypoint filtering. We showed that our proposed system performs better than the state of the art on challenging benchmark datasets. Further, the performance characterizations also revealed that the amount of visual information can be controlled using depth-based filtering to reduce perceptual aliasing, thereby leading to optimal performance under extreme appearance variations. In future, we plan to extend our work to a visual SLAM pipeline for robust metric localization and mapping under extreme appearance and viewpoint variations.

\bibliographystyle{IEEEtran}
\bibliography{newRefListCommon,depth}

\end{document}